\documentclass[letterpaper, 10 pt, conference]{ieeeconf}  % Comment this line out
\usepackage{amssymb}
\usepackage{amsmath}
\usepackage{graphicx}
\usepackage{graphics}
\usepackage{blkarray}
\usepackage{booktabs}
\usepackage{xcolor} 
\usepackage{bbold}
\usepackage{caption}
\usepackage{subcaption}
\usepackage{multirow}
\usepackage{comment}
\usepackage{epstopdf}

\usepackage[linesnumbered,ruled,vlined]{algorithm2e}

\usepackage{tikz}
\usepackage{todonotes}
\usetikzlibrary{arrows,matrix,positioning}
\graphicspath{{./figures/}}
\usepackage{caption}
\usepackage[labelformat=simple]{subcaption}
\usepackage{soul}
\usepackage{MnSymbol}
\usepackage{cite}

\DeclareCaptionLabelSeparator{periodspace}{.\quad}
\IEEEoverridecommandlockouts                              % This command is only
\newtheorem{remark}{Remark}
\newtheorem{assumption}{Assumption}

\newtheorem{problem}{Problem} 

\newtheorem{definition}{Definition}

\makeatletter
\newcommand*{\rom}[1]{\expandafter\@slowromancap\romannumeral #1@}
\makeatother

\newcommand{\subparagraph}{}

\newcommand{\qed}{\nobreak \ifvmode \relax \else
      \ifdim\lastskip<1.5em \hskip-\lastskip
      \hskip1.5em plus0em minus0.5em \fi \nobreak
      \vrule height0.75em width0.5em depth0.25em\fi}

\mathchardef\mhyphen="2D

\newcommand{\bt}{\mathcal{T}}

\title{\LARGE \bf
Learning of Behavior Trees for Autonomous Agents}

\author{Michele Colledanchise, Ramviyas Parasuraman, and Petter \"Ogren  % <-this % stops a space
\thanks{The authors are with the Centre for Autonomous Systems, Computer Vision and Active Perception Lab, School of Computer Science and Communication, The Royal Institute of Technology - KTH, Stockholm, Sweden.
e-mail: \tt{ $\{$miccol$|$ramviyas$|$petter$\}$@kth.se}}}

\begin{document}
\maketitle
\thispagestyle{empty}
\pagestyle{empty}
%%%%%%%%%%%%%%%%%%%%%%%%%%%%%%%%%%%%%%%%%%%%%%%%%%%%%%%%%%%%%%%%%%%%%%%%%%%%%%%%

\begin{abstract}

Definition of an accurate system model for Automated Planner (AP) is often impractical, especially for real-world problems. Conversely, off-the-shelf planners fail to scale up and are domain dependent. These drawbacks are inherited from conventional transition systems such as Finite State Machines (FSMs) that describes the action-plan execution generated by the AP. On the other hand, Behavior Trees (BTs) represent a valid alternative to FSMs presenting many advantages in terms of modularity, reactiveness, scalability and domain-independence.

In this paper, we propose a model-free AP framework using Genetic Programming (GP) to derive an optimal BT for an autonomous agent to achieve a given goal in unknown (but fully observable) environments. We illustrate the proposed framework using experiments conducted with an open source benchmark \textit{Mario AI} for automated generation of BTs that can play the game character \textit{Mario} to complete a certain level at various levels of difficulty to include enemies and obstacles.
%The use of Automated Planners (APs) in real-world problems is tricky. The definition of an accurate system model for planning is often impractical. On the other hand, off-the-shelf planners fail to scale up and are domain dependent. However these drawbacks are inherit from the transition system used to describe the action execution. Finite State Machines (FSMs) describe the transition system generated by the AP. Behavior Trees (BTs) represent a valid alternative to FSMs presenting many advantages in terms of modularity and reactiveness. We show how using BTs to describe the action execution allows us to improve the AP in terms of scalabilty and domain dependency. We provide a set of experimental results using a open source benchmark for testing APs.   
\end{abstract}


\section{Introduction}

Automated planning is a branch of Artificial Intelligence (AI) that concerns the realization of strategies or action sequences, typically for execution by intelligent agents, autonomous robots and unmanned vehicles. Unlike classical control and classification problems, the solutions are complex and must be discovered and optimized in multidimensional space. According to \cite{mitchell1997machine}, there are four common subjects that concern the use of Automated Planners (APs). First, the \emph{knowledge representation}. That is the type of knowledge that an AP will learn must be defined; Second, the \emph{extraction of experience}. That is how learning examples are collected; Third, the \emph{learning algorithm}. That is how to capture patterns from the collected experience; Finally, the \emph{exploitation of collected knowledge}. That is how the AP benefits from the learned knowledge.

Applying AP in a real word scenario is still an open problem \cite{jimenez2012}. In fully known environments with available models, the planning can be done offline. Solutions can be found and evaluated prior to execution. Unfortunately in most cases the environment is unknown and the strategy needs to be revised online. Recent works are extending the application of APs, from toy examples to real problems such as planning space mission \cite{nayak1999validating}, fire extinction, \cite{fdez2006bringing} underwater navigation \cite{bellingham2007robotics}. However, as highlighted in \cite{jimenez2012review}, most of these planners are hard to scale up and presents issues when it comes to extend their domain. 
Despite these successful examples, the application of APs to real worlds problem suffers of two main problems:
\emph{Planning Task}: Generally, APs require accurate description of the planning task. These descriptions include the model of the action that can be performed in the environment, the specification of the  state of the environment and the goal to achieve. Generating exact definition of the planning is often unfeasible for real-world problems;
\emph{Extensibility}: Usually, a solution of an AP is a PSpace-complete problem~\cite{bylander1994computational,bylander1991complexity}. Recent works tackle this problem through reachability analysis~\cite{bacchus2000using,nau2003shop2}, but still search control knowledge is more difficult than the planning task because it requires expertise in the task to solve as well as in the planning algorithm \cite{minton1988learning}. 
\begin{figure}[t]
\centering
\includegraphics[width=0.8\columnwidth]{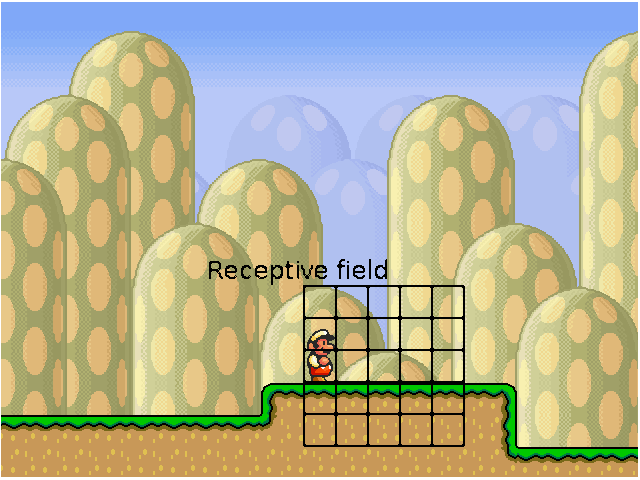}
\caption{Benchmark used to validate the framework.}
\label{IN.fig.mariofield}
\end{figure}

The task's goal is described using a fitness function defined by the user. The derived action execution is described as a composition of sub-planners using a tree-structured framework inherited from computer game industry~\cite{isla2008halo}, namely BT. BTs are a recent modular alternative to Controlled Hybrid Systems (CHSs) to describe reactive fault tolerant executions of robot tasks~\cite{colledanchise14}. BTs were first introduced in artificial intelligence for computer games, to meet their needs of programming the behavior of in-game non player opponents~\cite{ogren}. Their tree structure, which encompasses modularity; flexibility; and ease of human understanding, have made them very popular in industry, and their graph representations have created a growing amount of attention in academia~\cite{colledanchise14,colledanchise14-2,Bagnell2012b,klockner2013,Colledanchise15} and robotic industry~\cite{Kelleher15}. The main advantage of BTs as compared to CHSs can be seen by the following programming language analogy. In most CHSs, the state transitions are encoded in the states themselves, and switching from one state to the other leaves no memory of where the transition was made from. This is very general and flexible, but actually very similar to the now obsolete \emph{GOTO statement}, that was an important part of many early programming languages, e.g.,  BASIC. 

In BTs the equivalents of state transitions are governed by function calls and return values being passed up and down the tree structure. This is also flexible, but similar to the calls of FUNCTIONS that has replaced  GOTO in almost all modern programming languages. Thus, BTs exhibit many of the advantages in terms of readability, modularity and reusability that was gained when going from GOTO to FUNCTION calls in the 1980s. Moreover in a CHSs adding a state turns in evaluating each possible transition from/to the new state and removing a state can require the re-evaluation of all the transitions in the system. BTs reveal to have a natural way to connect/disconnect new states avoiding redundant evaluation of state transitions. In a tree-structured framework as BT, the relation between nodes are defined by parent-child relations. These relations are plausible in Genetic Programming (GP) allowing entire sub-trees to cross-over and mutate through generations to yield an optimized BT that generates a plan leading to the desired goal.

In this paper we propose a model free algorithm based framework that generates a BT for an autonomous agent to achieve a given goal in unknown environments. The advantages of our approach lies on the advantages of BTs over a general CHS. Hence our approach is modular and we can reduce the complexity dividing the goal in sub-goals.

\section{Related Work}
Evolutionary algorithms has been successfully applied in evolving robot or agent's behaviors \cite{parker2010,decroon2013,lazarus2001,perez2011evo}. For instance, in \cite{lazarus2001}, the authors used GP methodology to result in a better wall-follower algorithm for a mobile robot. In another interesting example by \cite{shanker2012}, the authors applied \emph{Grammatical Evolution} to generate different levels of simulation environment for a game benchmark (MarioAI). Learning the agent's behaviors using evolutionary algorithms has shown to outperform reinforcement learning strategies at least in agents that possess ambiguity in its perception abilities \cite{decroon2005}.

BTs are originally used in gaming industry where the computer (autonomous) player uses BTs for its decision making. Recently, there has been works to improve a BT using several learning techniques, for example, Q-learning \cite{dey2013} and evolutionary approaches \cite{perez2011evo,lim2010}.

In a work by Perez et. al. \cite{perez2011evo}, the authors used GE to evolve BTs to create a AI controller for an autonomous agent (game character). Despite being the most relevant work, we depart from their work by using a metaheuristic evolutionary learning algorithms instead of grammatical evolution as the GP algorithm provides a natural way of manipulating BTs and applying genetic operators. 

Scheper et. al \cite{scheper2014} applied evolutionary learning to BTs for a real-world robotic (Micro Air Vehicle) application. It appears as the first real-world robotic application of evolving BTs. They used a (sub-optimal) manually crafted BT as an initial BT in the evolutionary learning process, and conducted experiments with a flying robot, while the BT that controls the robot is learning itself in every experiment. Finally, they demonstrated significant improvement in the performances of the evolved final BT comparing to the initial user-defined BT. While we take inspirations from this work, the downside is that this work require an initial BT for it to work, which goes against our model-free objective.

Even though the above-mentioned works motivates our present research, we intend to use a model-free framework as against model-based frameworks or frameworks that needs extensive prior information. Hence, we propose a framework that is more robust and require no information about the environment but thrives on the fact that the environment is fully-observable. Although we do not make a direct comparison of our work with other relevant works in this paper, we envisage it in our further works.

\section{Background: BT and GP}
In this section we briefly describe BTs and GP. A more detailed description of BTs can be found in ~\cite{ogren}.

\subsection{Behavior Tree}

A Behavior Tree is a graphical modeling language and a representation for execution of actions based on conditions and observations in a system. While BTs have become popular for modeling the Artificial Intelligence in computer games, they are similar to a combination of hierarchical finite state machines or hierarchical task network planners. 

\begin{figure}[h]
\centering
\includegraphics[width=0.6\columnwidth]{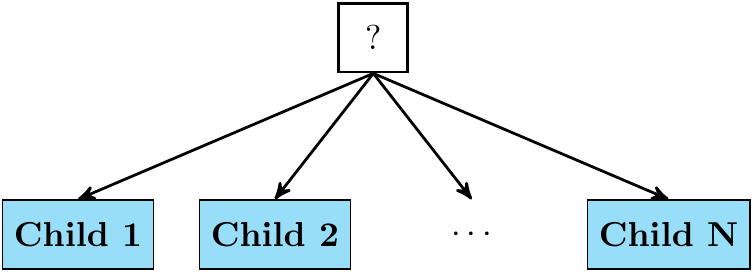}
\caption{Graphical representation of a fallback node with $N$ children.}
\label{bg.fig.sel}
\end{figure}

\begin{figure}[h]
\centering
\includegraphics[width=0.6\columnwidth]{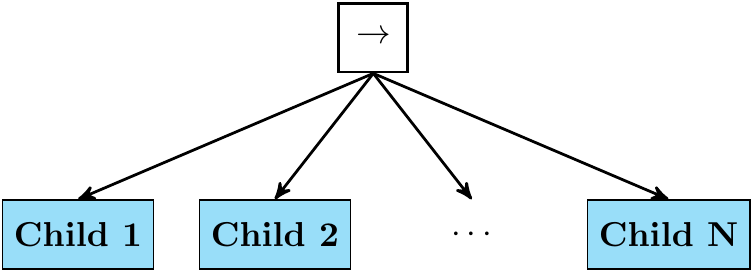}
\caption{Graphical representation of a sequence node with $N$ children.}
\label{bg.fig.seq}
\end{figure}

\begin{figure}[h]
\centering
\includegraphics[width=0.6\columnwidth]{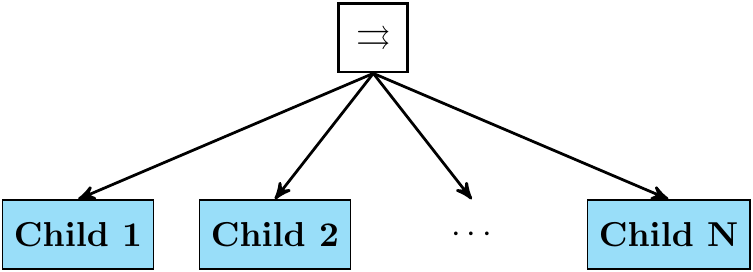}
\caption{Graphical representation of a parallel node with $N$ children.}
\label{bg.fig.par}
\end{figure}

A BT is a directed rooted tree where each node is either a control flow node or an execution node (or the root). For each connected nodes we define as \emph{parent} the outgoing node and \emph{child} the incoming node. The root has no parents and only one child, the control flow nodes have one parent and one or more child, and the execution nodes have one parent and no children. Graphically, the children of control flow nodes are placed below it. The children nodes are executed in the order from left to right, as shown in Fig.s~\ref{bg.fig.sel}-\ref{bg.fig.par}.

The execution of a BT begins from the root node. It sends \emph{ticks}~\footnote{A tick is a signal that allows the execution of a child} with a given frequency to its child. When a parent sends a tick to a child, the execution of this is allowed. The child returns to the parent a status \emph{running} if its execution has not finished yet, \emph{success} if it has achieved its goal, or \emph{failure} otherwise.\\ 
There are four types of control flow nodes (selector, sequence, parallel, and decorator) and two execution nodes (action and condition). Their execution is explained as follows.

\paragraph*{Selector}
The selector node ticks its children from the most left, returning success (running) as soon as it finds a child that returns success (running). It returns failure only if all the children return failure. When a child return running or success, the selector node does not tick the next children (if any).
The selector node is graphically represented by a box with a ``?", as in Fig.~\ref{bg.fig.sel}.
\begin{algorithm}[h]

  \For{$i \gets 1$ \KwSty{to} $N$}
  {
    \ArgSty{childStatus} $\gets$ \FuncSty{Tick(\ArgSty{child($i$)})}\\
    \uIf{\ArgSty{childStatus} $=$ \ArgSty{running}}
    {
      \Return{running}
    }
    \ElseIf{\ArgSty{childStatus} $=$ \ArgSty{success}}
    {
      \Return{success}
    }
  }
  \Return{failure}
  \caption{Pseudocode of a fallback node with $N$ children}
  
\end{algorithm}

\paragraph*{Sequence}
The sequence node ticks its children from the most left, returning failure (running) as soon as it finds a child that returns failure (running). It returns success only if all the children return success. When a child return running or failure, the sequence node does not tick the next children (if any). The sequence node is graphically represented by a box with a ``$\rightarrow$", as in Fig.~\ref{bg.fig.seq}.
\begin{algorithm}[h]

  \For{$i \gets 1$ \KwSty{to} $N$}
  {
    \ArgSty{childStatus} $\gets$ \FuncSty{Tick(\ArgSty{child($i$)})}\\
    \uIf{\ArgSty{childStatus} $=$ \ArgSty{running}}
    {
      \Return{running}
    }
    \ElseIf{\ArgSty{childStatus} $=$ \ArgSty{failure}}
    {
      \Return{failure}
    }
  }
  \Return{success}
  \caption{Pseudocode of a sequence node with $N$ children}
  
\end{algorithm}

\paragraph*{Parallel}
The parallel node ticks its children in parallel and returns success if $M \leq N$ children return success, it returns failure if $N-M+1$ children return failure, and it returns running otherwise.
The parallel node is graphically represented by a box with two arrows, as in Fig.~\ref{bg.fig.par}.
\begin{figure}[h]
        \centering
        \begin{subfigure}[b]{0.3\columnwidth}
                \centering
                \includegraphics[width=0.5\columnwidth]{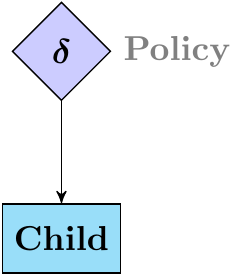}
                \caption{Decorator node. The label describes the user defined policy.}
                \label{bg.fig.dec}
        \end{subfigure}%
       ~ %add desired spacing between images, e. g. ~, \quad, \qquad etc.
        \begin{subfigure}[b]{0.3\columnwidth}
                \centering
                \includegraphics[width=0.3\columnwidth]{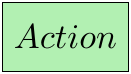}
                \caption{Action node. The label describes the action performed}
                \label{bg.fig.act}              
        \end{subfigure}
        ~ %add desired spacing between images, e. g. ~, \quad, \qquad etc.
        \begin{subfigure}[b]{0.3\columnwidth}
                \centering
                \includegraphics[width=0.5\columnwidth]{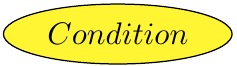}
                \caption{Condition node. The label describes the condition verified}
                \label{bg.fig.cond}
        \end{subfigure}
        \caption{Graphical representation of a decorator, action, and condition node.}
\end{figure}
\paragraph*{Decorator} 
The decorator node manipulates the return status of its child
 according to the policy defined by the user (e.g. it inverts the success/failure status of the child). The  decorator is graphically represented in Fig.~\ref{bg.fig.dec}.
\paragraph*{Action}
The action node performs an action, returning success if the action is completed and failure if the action cannot be completed. Otherwise it returns
 running.
  The action node is represented in Fig.~\ref{bg.fig.act} 
\paragraph*{Condition}
The condition node check whenever a condition is satisfied or not, returning success or failure accordingly.  The condition node never returns running. The condition node is represented in Fig.~\ref{bg.fig.cond} 
\paragraph*{Root}
The root node generates ticks. It is graphically represented as a white box labeled with ``$\varnothing$''

\subsection{Genetic Programming}

GP is an optimization algorithm, takes inspiration from biological evolution techniques and is a specialization of genetic algorithms where each individual itself is a computer program (in this work, each individual is a BT). We use GP to optimize a \emph{population} of randomly-generated BT according to a user-defined \emph{fitness function} determined by a BT's ability to achieve a given goal.

GP has been used as a powerful tool to solve complex engineering problems through evolution strategies \cite{rechenberg1994evolution}. In GP individuals are BTs that are evolved using genetic operations of reproduction, cross-over, and mutation. In each \emph{population}, individuals are selected according to the \emph{fitness function} and then mated, crossing over parts of their sub-trees to form an \emph{offspring}. The offspring is finally mutated generating a new population. This process continues until the GP finds a BT that satisfies the goal (such as minimize the fitness function and satisfy all constraints) is reached.

Often, the size of the final generated BT using GP is large even though there might exist smaller BT with the same fitness value and performance. This phenomenon of generating a BT of larger size than necessary can be termed as bloat. We also apply boat control at the end to optimize the size of the generated BT.

The GP used with BTs allows entire sub-trees to cross-over and mutate through generations. The previous generation are called parents and produces children BTs after applying genetic operators. The best performing children are selected from the child population to act as the parent population for the next generation. 

Crossover, mutation and selection are the three major genetic operations that we use in our approach. The crossover performs exchanges of subtrees between two parent BTs. Mutation is an operation that replace a node with a randomly selected parent BT. Selection is the process of choosing BTs for the next population. The probability of being selected for the next population is proportional to a \emph{fitness} function that describes ``how close" the agent is from the goal.

\subsubsection{Two-point crossover in two BTs}
\begin{figure}[h]
\centering
\includegraphics[width=0.8\columnwidth]{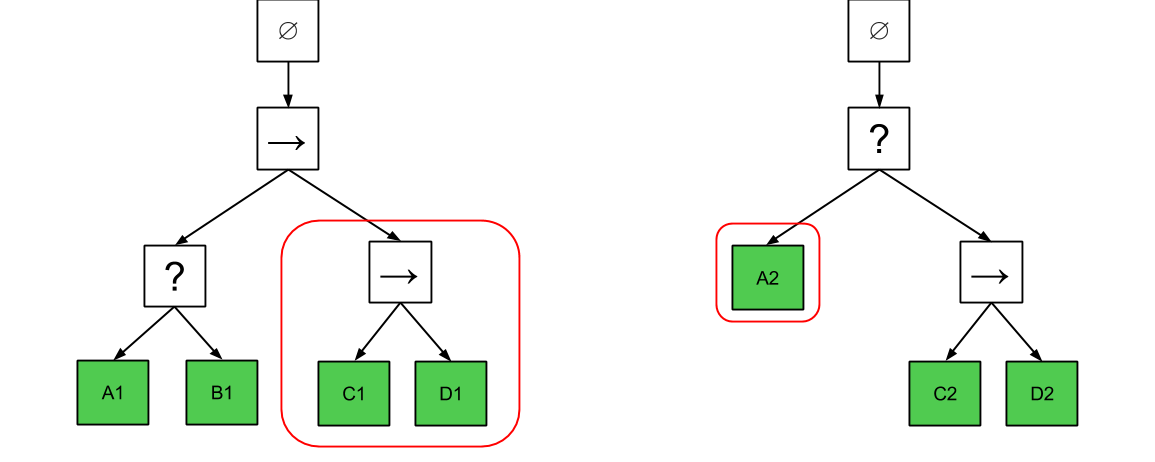}
\caption{BTs before the cross over of the highlighted sub-trees.}
\label{PA.fig.CrossoverBefore}
\end{figure}

\begin{figure}[h]
\centering
\includegraphics[width=0.8\columnwidth]{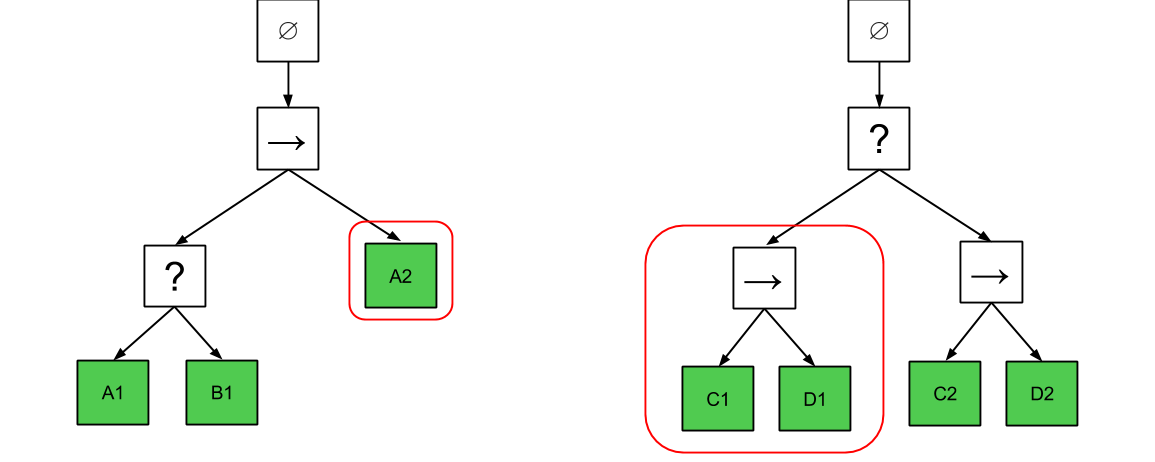}
\caption{BTs after the cross over of the highlighted sub-trees.}
\label{PA.fig.CrossoverAfter}
\end{figure}

The crossover is performed by randomly swapping a sub-tree from a BT with a sub-tree of another BT at any level \cite{tweedale2012innovation}. Fig.~\ref{PA.fig.CrossoverBefore} and Fig.~\ref{PA.fig.CrossoverAfter} show two BTs before and after a cross over operation.

\begin{remark}
Note that, using BTs as the knowledge representation framework, avoids the problem of logic violation during cross-over experienced in \cite{fu2003genetic}.
\end{remark}

\subsubsection{Mutation operation in BT}
We use unary mutation operator where the mutation is carried out by replacing a node in a BT with another node of the same type (i.e. we do not replace a execution node with a control flow node or vice versa). This increases diversity, crucial in GP \cite{tweedale2012innovation}. To improve convergence properties we use the so-called \emph{simulated annealing}~\cite{Davis1987} performing the mutation on several nodes of the first population of BT and reducing gradually the number of mutated nodes in each new population. In this way we start with a very high diversity to avoid possible local minima of the fitness function and we get a smaller diversity as we get close to the goal.

\subsubsection{Selection mechanism}
From the mutated population, also called \emph{offspring}, individuals are selected for the next population. The selection process is a random process which select a given $\bt_i$ with a probability $p_i$. This probability is proportional to the a \emph{fitness function} $f_i$ which quantitatively measures how many sub-goals are satisfied. There are three most used way to compute $p_i$ given the fitness function \cite{mitchell1997machine}:

\begin{enumerate}
\item Naive Method: $p_i=\frac{f_i}{\sum_j f_j}$ that is the fitness divided by the sum of all the fitness of the individuals in the population (to ensure $p_i\in[0,1]$)
\item Rank Space Method: We $P_c$ set as the probability of the highest ranking individual (individual with highest $f_i$), then we sort the trees in the population in descending order w.r.t. the fitness. (i.e. $\bt_i$ has higher or equal fitness of $\bt{i-1}$). Then then the probabilities $p_i$ are defined as follows.
\begin{eqnarray}
p_k = &(1-P_c)^{k-1}P_c \; \forall k \in \{1,2,\ldots,N-1 \} \\
p_N = &(1-P_c)^{k-1}
\end{eqnarray}

%\begin{figure}[h]
%\centering
%\includegraphics[width=0.8\columnwidth]{curves.eps}
%\caption{Iso-goodnes curves with $d_i=10$ and $f_i=10$.}
%\label{PS.fig.isogood}
%\end{figure}
\item Diversity Rank Method: We measure the \emph{diversity} $d_i$ of an individual $\bt_i$ w.r.t. the others in the population. The probability $p_i$ encompasses both diversity and fitness. Let $\bar d$ and $\bar f$ be the maximal value of $d_i$ and $f_i$ in the population respectively, the probability $p_i$ is given by:
\begin{equation}
p_i = 1- \frac{||[d_i,p_i] - [\bar d,\bar p]||}{||[\bar d,\bar p]||}
\end{equation}
Individuals with the same survival probability lies on the so-called \emph{iso-goodness} curves.
\end{enumerate}

\section{Problem Formulation}
Here we formulate definitions and assumptions, then we state the main problem, and finally illustrate the approach with an example.

\begin{assumption}
The environment is unknown but fully observable. We consider the problem so called \emph{learning stochastic models in fully observable environment} \cite{jimenez2012}. 
\end{assumption}

\begin{assumption}
There exists a finite set of actions that, performed, lead from the initial condition to the goal.
\end{assumption}

\begin{definition}
$\mathcal{S}$ is state space of the environment.
\end{definition}
\begin{remark}
We only know the initial state and the final state.
\end{remark}
\begin{definition}
$\Sigma$ is a finite set of actions.
\end{definition}

\begin{definition}
$\gamma:\mathcal{S} \times \Sigma \to [0,1]$ is the \emph{fitness} function. It takes value $1$ if and only if performing the finite set of actions $\Sigma$ change the state of the environment from an initial state to a final state that satisfies the goal.
\end{definition}

\begin{problem}
\label{PF.p1}
Given a goal described by a fitness function $\gamma$ and an arbitrary initial state $s_0\in \mathcal{S}$ derive an action sequence $\Sigma$ such that $\gamma(s_0,\Sigma) = 1 $. 
\end{problem}

\section{Proposed approach}

In this section we describe the proposed approach. We begin with defining which actions the agent can perform and which conditions it can observe. We also define the appropriate fitness function that takes input a BT and results in a fitness value proportionate to how closer the BT is in achieving a given goal. An empirically determined moving time widow $\tau$ (seconds) is used in the execution process, where the BT is executed continuously but the fitness function is evaluated for the past $\tau$ seconds. The progressive change in fitness function are assessed to determine the course of the learning algorithm. 

We follow a metaheuristic learning strategy, where we use a greedy algorithm first and when it fails, we use the GP. The GP will also be used when the greedy algorithm cannot provide any results or when the complexity of the solution is increased. This mixed-learning based heuristic approach was to minimize the learning time significantly compared to using pure GP, while still achieving an optimal BT that satisfies a given goal.

At an initial state $s_0$ we start with a BT that consists of only one node which is an action node. To choose that action node, we use a greedy search process where each action is executed until we find an action that, when executed for the $\tau$ seconds, the value of the fitness function keep increasing. if such an action is found, that action is added to the BT. However if no actions are found, the GP process will be initiated with a population of binary trees (two nodes in a BT) with random node assignments consisting of combination of condition and action nodes, and results in an initial BT that increases the fitness value the most.

In the next stages, the resultant BT is executed again, and the changes in the conditions and fitness values are monitored. When the fitness value starts decreasing, the recently changed conditions (within $\tau$ seconds) will be composed (randomly) as a subtree (as in Fig.~\ref{bg.fig.sel} and added to the existing BT. Then, we use the greedy search algorithm as above to find the action node for the previously added subtree by adding each possible actions to that subtree and whole BT is executed.  Once again, when an action that increase the fitness value is found, that action is added to the recent subtree of the BT and the whole process continues. If no such action is found then the GP will be used to determine the subtree which increases the fitness value. We iterate these processes until the goal is achieved. Finally, we remove the possible unnecessary nodes in the BT by applying anti-bloat control operation. 

We now address the concerns raised by~\cite{mitchell1997machine} using the AP proposed using BTs.
\subsection{Knowledge representation}
The knowledge is represented as a BT. A BT can be seen as a rule-based system where it describes which action to perform when some conditions are satisfied.

\subsection{Extraction of the experience}
%For every new initial state $s_0$ the framework learns a new BT. We make fallback composition of learned BTs. For each $s_0 \in \mathcal{S}$ the BT runs the related sub tree.
The experiences (knowledge) are extracted in terms of conditions observed from the environment using sensory perceptions in the autonomous agents. Examples of conditions for a robot could be obstacle position, position information, energy level, etc. Similarly, example conditions for a game character could be "enemy in front", "obstacle close-by", "level reached", "points collected", "number of bullets remaining", etc.

\subsection{Learning algorithm}
\label{PA.LA}
Algorithm~\ref{PS.ALG.LA} presents the pseudo-code for the learning algorithm. The learning algorithm has $2$ steps. The first step aims to identify which condition have to be verified in order to perform some actions. The second step aims to learn the actions to perform.

As mentioned earlier, the framework starts at the initial state $s_0$. If the value of the fitness function does not increase, a greedy algorithm is used to try each action until it finds the one that leads to an increase of the fitness value. If no actions are found, it start the GP to learn a BT composition of actions as explained before. We call the learned BT $\bt_0$.
\begin{remark}
In case the framework learns a single action, $\bt_0$ is a degenerate BT composed by a single action.
\end{remark} 
%The framework runs the learned BT, if any, as long as the fitness function increases. During this time the framework monitors the values of the conditions. When the fitness stops to increase, the conditions that have changed in a time window $\tau$ are stored in a BT composition. The user can choose the time window. Now, we need to create a BT composition of those conditions. The BT composition returns success only when the conditions changed during $\tau$ return the corresponding value i.e. the conditions that have changed to true returns success and the conditions that have changed to false returns failure. 

\begin{figure}[h]
\centering
\includegraphics[width=0.8\columnwidth]{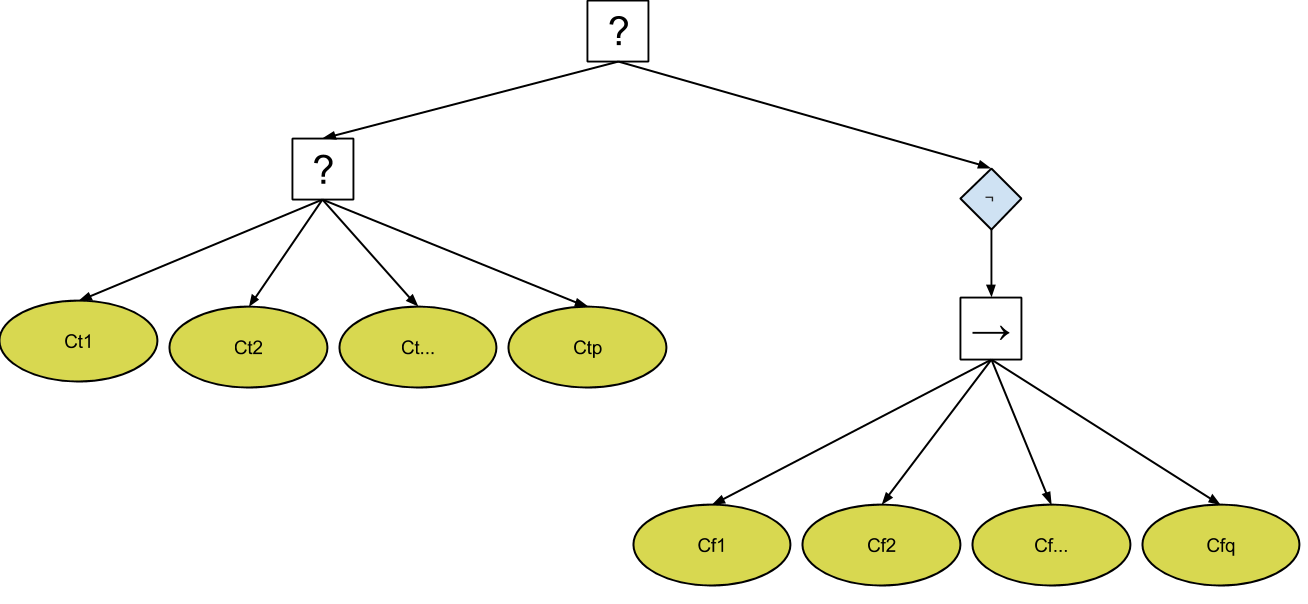}
\caption{Graphical representation of $\bt_{cond}$.}
\label{PA.fig.BTcond}
\end{figure}

Let $C_F \subseteq \mathcal{C}$ be the set of conditions that have changed from true to false and let $C_T \subseteq \mathcal{C}$ be the set of conditions that have changed from false to true during $\tau$. The BT composition of those conditions, $\bt_{cond}$, is depicted in Fig.~\ref{PA.fig.BTcond}.

The conditions encoded in $\bt_{cond1}$ make the fitness value decrease (i.e. when $\bt_{cond1}$ return true, the fitness value decreases). Thus we need to learn the BT to be performed whenever $\bt_{cond1}$ returns success to enable increase in fitness value. The learning procedure continues. Let $\bt_{acts1}$ be the learned BT to be performed when $\bt_{cond1}$ returns success, the BT that the agent runs is now 
\begin{equation}
\bt_1\triangleq \mbox{selector}(\tilde\bt_1, \bt_0)
\end{equation}
where  
\begin{equation}
\tilde\bt_1\triangleq \mbox{selector}(\bt_{cond1}, \bt_{acts1}).
\end{equation}
The agent runs $\bt_1$ as long as the value fitness function increases. When the fitness stops to increase a new BT is learned following the previous procedure $\bt_1$. Generally as long as the goal is not reached, the learned BT is:
 \begin{equation}
\bt_{i}\triangleq \mbox{selector}(\tilde\bt_i, \bt_{i-1})
\end{equation} 

When the final BT is learned (i.e. that BT that leads the agent to the goal), we run the anti-bloat algorithm to remove possibly inefficient nodes introduced due to a large time window $\tau$ or due to the randomness needed for the GP. 

\begin{algorithm}[h]

  \ArgSty{$\gamma_{old}$} $\gets$ \FuncSty{GetFitness(\ArgSty{nil})}\\
  \ArgSty{$t_0$} $\gets$ \FuncSty{GetFirstBT()}\\
    \ArgSty{$t$} $\gets$ \ArgSty{$t_0$} \\

  \While{$\gamma < 1$}
  {
	  \ArgSty{$\gamma$} $\gets$ \FuncSty{GetFitness(\ArgSty{$t$})}\\
	  
    \If{\ArgSty{$\gamma_{old}$} $\geq$ \ArgSty{$\gamma$}}
    {
  		\ArgSty{$t_{cond}$} $\gets$ \FuncSty{GetChangedConditions()}\\
  		\ArgSty{$t_{acts}$} $\gets$ \FuncSty{LearnSingleAction(\ArgSty{$t$})}\\
		\If{\ArgSty{$t_{acts}$} $=$ \ArgSty{nil} }
		    {
			\ArgSty{$t_{act}$} $\gets$ \FuncSty{LearnBT(\ArgSty{$t$})}\\
		    }
		\ArgSty{$\tilde t$} $\gets$ \FuncSty{Sequence(\ArgSty{$t_{cond}$},\ArgSty{$t_{acts}$})}\\

    }
	    \ArgSty{$t$} $\gets$ \FuncSty{Selector(\ArgSty{$\tilde t$},\ArgSty{$t$})}

  \ArgSty{$\gamma_{old}$} $\gets$ \ArgSty{$\gamma$}\\
  }
  \Return{$t$}\\
  \caption{Pseudocode of the learning algorithm}
  \label{PS.ALG.LA}
  
\end{algorithm}

\subsection{Exploitation of the collected knowledge}
At each stage, the resulting BTs are executed in a simulated (or real) environment to evaluate against a fitness function. Based on the value of the fitness function, the learning algorithm decides the future course. The fitness function is defined in accordance to a given goal. For instance, if the goal is to complete a level in a game, then the fitness function is a function of the following: the game points acquired by the agent (game character), how far (distance) the agent has traversed, how much time the agent has spent in the game level, how many enemies are shot by the agent, etc.
%\subsection{Genetic Programming}

\subsection{Anti-bloat control}

Once we obtained the BT that satisfies the goal, we search for ineffective sub-trees, i.e. those action compositions that are superfluous for the goal reaching. This process is called anti-bloat control in GP. Most often, the genetic operators (such as size fair crossover and size fair mutation) or the selection mechanism in GP applies the bloat control. However, in this work, we first generate the BT using the GP without size/depth restrictions in order to achieve a complex yet practical BT. Then we apply bloat control using a separate breadth-first algorithm that reduces the size and depth of the generated BT while keeping the properties of the BT and its performance at the same time.

To identify the redundant or unnecessary sub-trees, we enumerate the sub-trees with a Breadth-first enumeration. We run the BT without the first sub tree and we check if the fitness function has a lower value or not. In the former case the sub-tree is kept, in the latter case the sub-tree is removed creating a new BT without the sub-tree mentioned. Then we run the same procedure on the new BT. The procedure stops when there are no ineffective sub-tree found. Algorithm~\ref{PS.ALG.REM} presents a preudo-code of the procedure.

\begin{algorithm}[h]

    \ArgSty{$t_{new}$} $\gets$  \ArgSty{t} \\
    \ArgSty{i} $\gets$  0 \\

  \While{$i \leq$  \FuncSty{GetNodesNumber(\ArgSty{$t_{new}$})}}
  {
    \ArgSty{i} $\gets$  \ArgSty{i} + 1 \\
    \ArgSty{$t_{rem}$} $\gets$ \FuncSty{RemoveSubtree(\ArgSty{$t_{new}$},$i$)}\\
    \If{\FuncSty{GetFintess(\ArgSty{$t_{rem}$})} $\geq$ \FuncSty{GetFintess(\ArgSty{$t_{new}$})}}
    {
              \ArgSty{$t_{new}$} $\gets$  \ArgSty{$t_{rem}$} \\
               \ArgSty{i} $\gets$  0 \\
    }
  }
  \Return{\ArgSty{$t_{new}$}}
  \caption{Pseudocode of a anti-bloat control for inefficient subtree(s) removal.}
  \label{PS.ALG.REM}
\end{algorithm}

\begin{remark}
The procedure is trivial using a BT due to its tree structure.   
\end{remark}

\section{Preliminary experiments}

To experimentally verify the proposed approach, we used the Mario AI \cite{karakovskiy2012mario} open-source benchmark for the Super Mario Bros game developed initially by Nintendo. The gameplay in Mario AI, as in the original Nintendo's version, consists in moving the controlled character, namely Mario, through two-dimensional levels, which are viewed sideways. Mario can walk and run to the right and left, jump, and (depending on which state he is in) shoot fireballs. Gravity acts on Mario, making it necessary to jump over cliffs to get past them. Mario can be in one of three states: \emph{Small}, \emph{Big} (can kill enemies by jumping onto them), and \emph{Fire} (can shoot fireballs).

The main goal of each level is to get to the end of the level, which means traversing it from left to right. Auxiliary goals include collecting as many coins as possible, finishing the level as fast as possible, and collecting the highest score, which in part depends on number of collected coins and killed enemies.

Complicating matters is the presence of cliffs and moving enemies. If Mario falls down a hole, he loses a life. If he touches an enemy, he gets hurt; this means losing a life if he is currently in the Small state. Otherwise, his state degrades from Fire to Big or from Big to Small. 

\paragraph{Actions}
In the benchmark there are five action available: Walk right, walk left, crouch, shoot, and jump.
\paragraph{Conditions}

\begin{figure}[h]
\centering
\includegraphics[width=0.8\columnwidth]{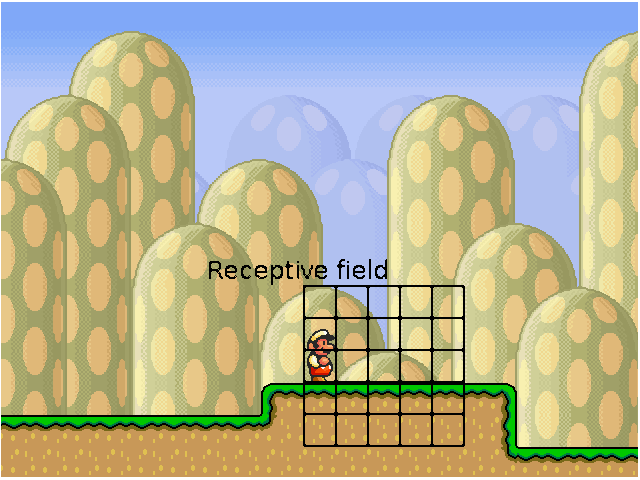}
\caption{Receptive field around Mario. In this case Mario is in state Fire, hence he occupies 2 blocks.}
\label{ER.fig.mariofield}
\end{figure}

In the benchmark there is a receptive field of observations. We chose a $5 \times 5$ grid for such receptive field as shown in Fig.~\ref{ER.fig.mariofield}. For each box of the grid there are 2 conditions available: if the box is occupied by an enemy and if the box is occupied by an obstacle. For a total of $50$ conditions.

\paragraph{Fitness Functions}
The fitness function is given by a non linear combination of the distance passed, enemy killed, number of hurts, and time left when the end of the level is reached.
Fig.\ref{ER.fig.mariofield} illustrates the receptive field around Mario, used our experiments.

\subsection{Testbed 1: No enemies and no cliffs}
\begin{figure}[h]
\centering
\includegraphics[width=0.8\columnwidth]{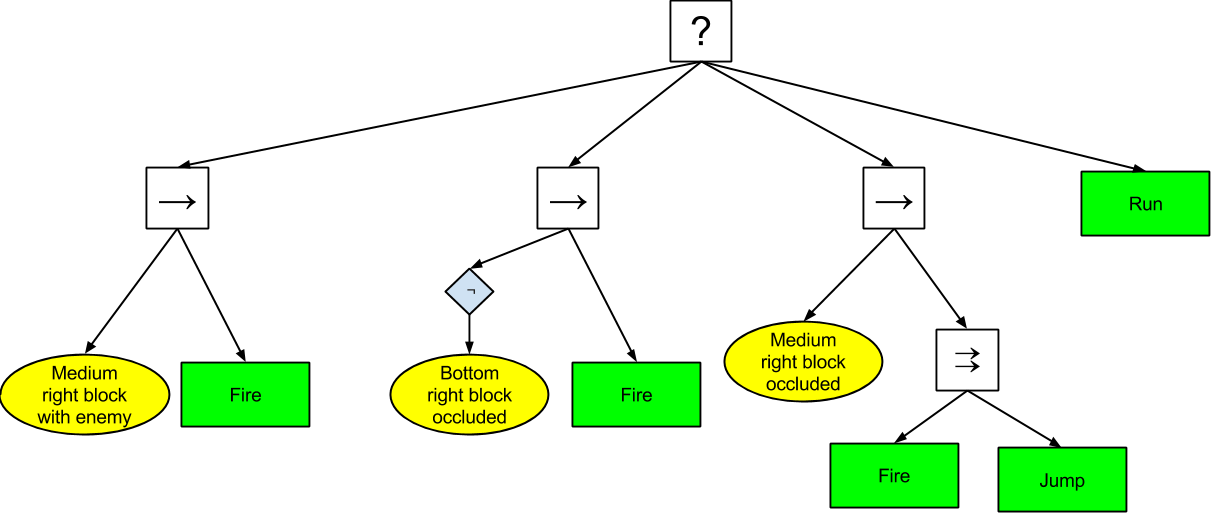}
\caption{BT learned for Testbed 1.}

\label{ER.fig.marioAIBT}
\end{figure}

This is a simple case. The agent has to learn how to move towards the end of the level and how to jump obstacles. The selection method in the GP is the rank-space method.
A youtube video shows the learning phase in real time (https://youtu.be/uaqHbzRbqrk).
Fig. \ref{ER.fig.marioAIBT} illustrates a resulting BT learned for the Testbed 1.

\subsection{Testbed 2: Walking Enemies and No Cliffs}

\begin{figure}[h]
\centering
\includegraphics[width=0.8\columnwidth]{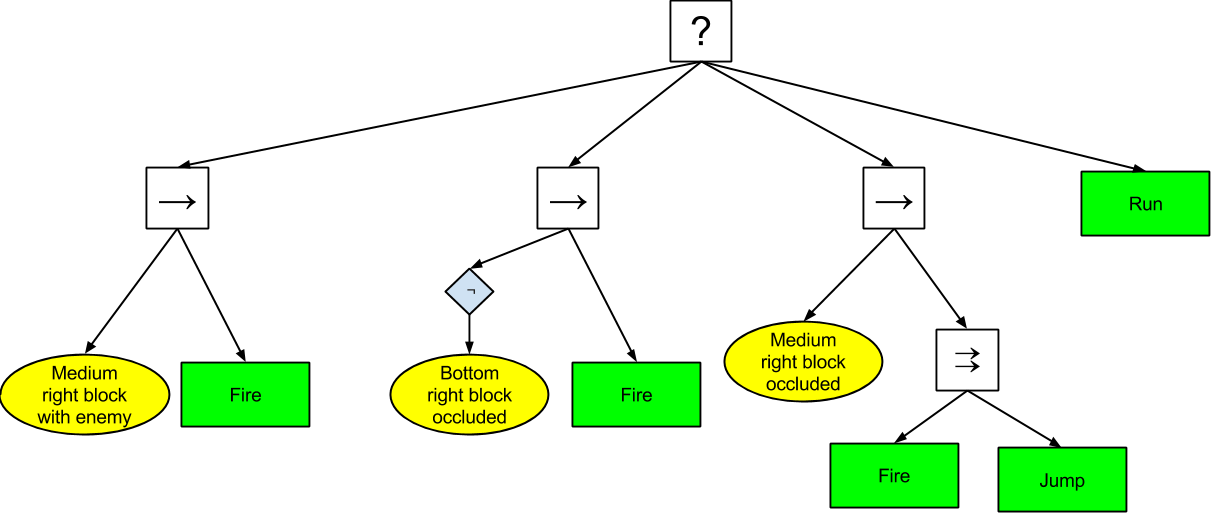}
\caption{BT learned for Testbed 2.}
\label{ER.fig.marioAIBT2}
\end{figure}

This is slightly more complex than Tesbed 1. The agent has to learn how to move towards the end of the level, how to jump obstacles, and how to kill the enemies. The selection method in the GP is the rank-space method.
A youtube video shows the learning phase in real time (https://youtu.be/phy98jbdgQc).
\begin{remark}
The youtube video does not show the initial BT learned $\bt_0$, which was a simple action ''Right".
\end{remark}
Fig. \ref{ER.fig.marioAIBT2} illustrates a resulting BT learned for the Testbed 2.

\subsection{Testbed 3: Flying Enemies and Cliffs}
For this testbed we used a newer version of the Benchmark. The agent has to learn how to move towards the end of the level, how to jump obstacles, and how to kill the enemies and how to avoid cliffs. The enemies in this testbed can fly. The selection method in the GP is the rank-space method.
A youtube video shows the final BT (https://www.youtube.com/watch?v=YfvdHY-DXwM).

We avoid to depict the final generated BT because of size restrictions (about 20 nodes).

\section{Conclusions}
BT are used to represent the knowledge as it provides a valid alternative to conventional planners such as FSM, in terms of modularity, reactiveness, scalability and domain-independence. In this paper we presented a model-free AP for an autonomous agent using metaheuristic optimization approach involving a combination of GP and greedy-based algorithms to generate an optimal BT to achieve a desired goal. To our best knowledge, this is the first work following a fully model-free framework whereas other relevant works either use model based frameworks or use apriori information for the behavior trees. We have detailed how we addressed the following subjects in AP: knowledge representation; learning algorithm; extraction of experience and exploitation of the collected knowledge. Further, the proposed approach was tested in the open-source ``Mario AI" benchmark to simulated autonomous behavior of the game character "Mario" in the benchmark simulator. Some samples of the results are illustrated in this paper. A video of an working example and illustration is available at https://youtu.be/phy98jbdgQc. Even though the results are encouraging and comparable to the state-of-the-art, more vigorous analysis and validation will be needed before extending the proposed approach to real-world robots.

\section{Future Work}
The first future work is to examine our approach in the Mario AI benchmark with extensive experiments and compare our results with other state-of-the-art approaches such as \cite{perez2011evo}. We further plan to explore dynamic environments and adapt our algorithm accordingly. Inspired by the work in \cite{tomai2014}, we also plan to look at the possibility of using supervised learning to generate an optimal BT.

Regarding the supervised learning, we are developing a model-free framework to generate BT by learning from training examples. The strength of the approach lies on the possibility of separating the tasks to learn. A youtube video (http://youtu.be/ZositEzjidE) shows a preliminary result of the supervised learning approach implemented in the MarioAI benchmark. In this example, the agent learns separately the task \emph{shoot} and the task \emph{jump} from examples of a game played by an user.

\bibliographystyle{IEEEtran}
\bibliography{EvolBT}	
\end{document}